\documentclass[runningheads]{llncs}

\usepackage[T1]{fontenc}
\usepackage{graphicx}
\usepackage[colorinlistoftodos,textsize=tiny]{todonotes}
\usepackage{xspace}

\usepackage{multirow}

\usepackage{caption}
\usepackage{subcaption}

\usepackage{url}

\usepackage{amsmath}

\newcommand{\dbname}{LSA-T\xspace}

\newcommand{\dbrepourl}{\url{https://github.com/midusi/LSA-T}}
\newcommand{\modelrepourl}{\url{https://github.com/midusi/keypoint-models}}

\usepackage{hyperref}
\begin{document}

\title{\dbname: The first continuous Argentinian Sign Language dataset for Sign Language Translation}
\titlerunning{\dbname: The first continuous LSA dataset}

\author{
Pedro Dal Bianco\inst{1,3}\orcidID{0000-0001-7197-8602} \and
Gastón Ríos\inst{1,3}\orcidID{0000-0003-0252-7036} \and
Franco Ronchetti\inst{1,2}\orcidID{0000-0003-3173-1327} \and
Facundo Quiroga\inst{1,4}\orcidID{0000-0003-4495-4327} \and
Oscar Stanchi\inst{1}\orcidID{0000-0003-0294-2053} \and
Waldo Hasperué\inst{1,2}\orcidID{0000-0002-9950-1563} \and
Alejandro Rosete \inst{5} \orcidID{0000-0002-4579-3556}
}
\authorrunning{P. Dal Bianco et al.}
%
\institute{Instituto de Investigación en Informática LIDI - Universidad Nacional de La Plata. Argentina \and
Comisión de Investigaciones Científicas de la Pcia. de Bs. As. (CIC-PBA). Argentina \and
Becario Doctoral UNLP \and Becario Postdoctoral UNLP \and
Universidad Tecnológica de La Habana José Antonio Echeverría. Cuba.
}
\maketitle              
\begin{abstract}
Sign language translation (SLT) is an active field of study that encompasses human-computer interaction, computer vision, natural language processing and machine learning. Progress on this field could lead to higher levels of integration of deaf people. This paper presents, to the best of our knowledge, the first continuous Argentinian Sign Language (LSA) dataset. It contains 14,880 sentence level videos of LSA extracted from the CN Sordos YouTube channel with labels and keypoints annotations for each signer. We also present a method for inferring the active signer, a detailed analysis of the characteristics of the dataset, a visualization tool to explore the dataset and a neural SLT model to serve as baseline for future experiments.

\end{abstract}

\section{Introduction}
\label{sec:intro}
Sign language is one of the main tools that the deaf community has both to communicate and to access information. Sign Language Recognition (SLR) has been an active research field for the last two decades \cite{camgoz2018neural}. However, each sign language has its own specific linguistic rules \cite{stokoe1980sign} and there is no direct correspondence between a sequence of words in a traditional spoken language and a sequence of signs. Therefore, an approach based on SLR, as a special case of a gesture recognition problem, is not enough to close the gap between deaf and hearing people. The problem of translating directly from sign language videos to text in the corresponding language is known as Sign Language Translation (SLT).
 Compared to other translation tasks, SLT presents unique challenges \cite{camgoz2018neural}:

\begin{itemize}

    \item The signs used and their meanings vary in different countries and even in different regions, meaning that for a model to be able to translate a specific sign language, for example, Argentinian Sign Language (LSA), it needs to be trained on data from that specific language.
    
    \item A model for SLT needs to learn not only the meaning of each sign but also the temporal dynamics and dependences between signs. Therefore, it has to be trained on videos that contain sequences of (continuous sign language or CSL). Datasets that contain isolated signs (videos containing one sign each) are not suitable for SLT.
    
    \item There are much fewer labeled sign language videos than in other domains (such as audio transcription) and there are also relatively few sign language users in the general population, which makes it difficult to find interpreters to label videos.
    
    \item Each sign is made of many components such as hand shapes, positions, movements, body poses, facial expressions, and lip movements. A model able to perform SLT should be able to use all these features.
    
    \item Typically, available videos of sign language share the same backgrounds, interpreters, or topics. This is specially true for laboratory-made datasets or, for example, videos from news channels that have a live interpreter. Models trained on this data might be biased toward these features and unable to generalize to other people or backgrounds.
    
\end{itemize}
\begin{figure}[!t]
     \centering
     \begin{subfigure}[b]{0.47\textwidth}
         \centering
         \includegraphics[width=\textwidth]{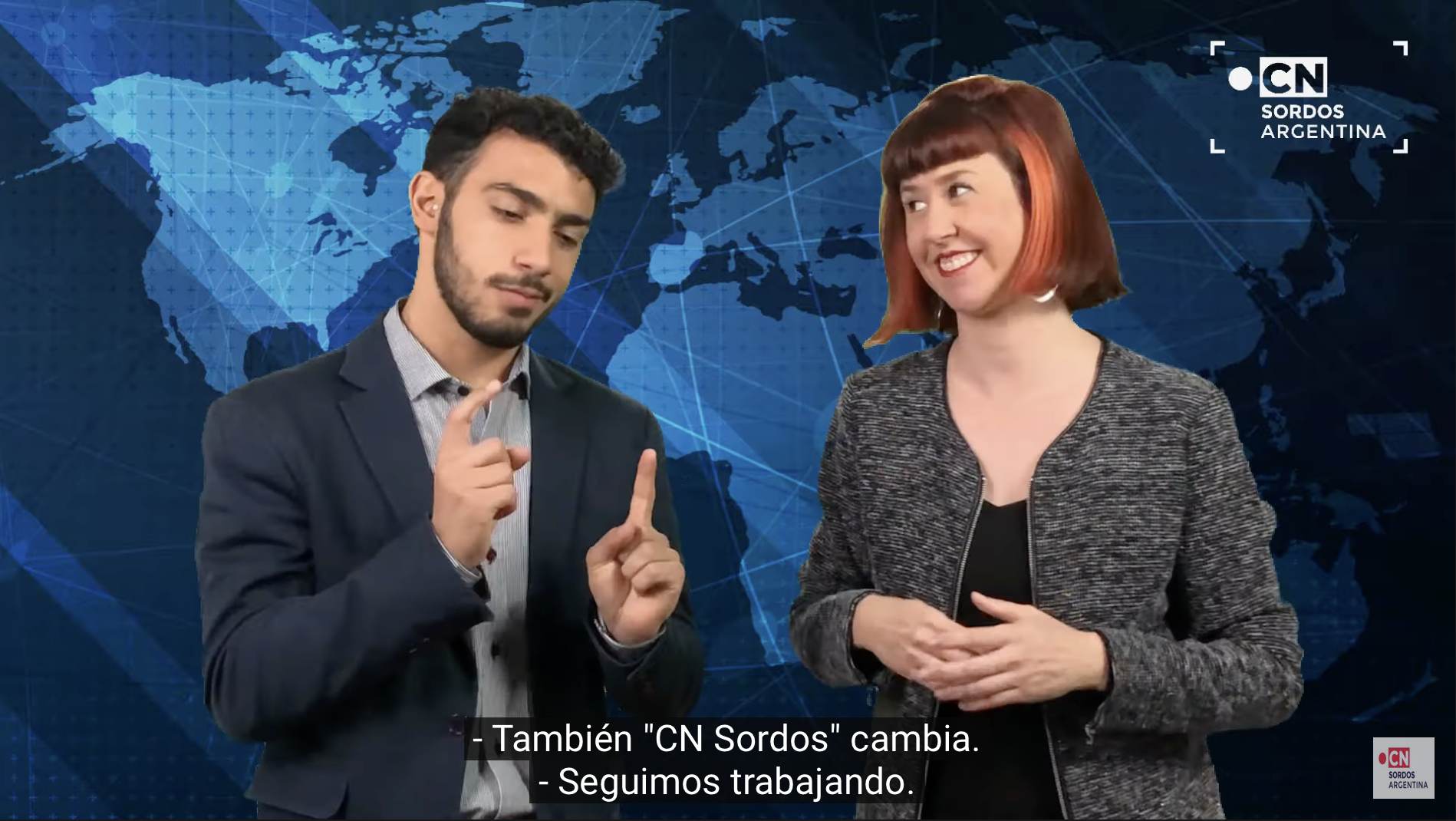}
         \caption{"Resumen semanal"}
         \label{subfig:resumen_semanal}
     \end{subfigure}
     \hfill
     \begin{subfigure}[b]{0.47\textwidth}
         \centering
         \includegraphics[width=\textwidth]{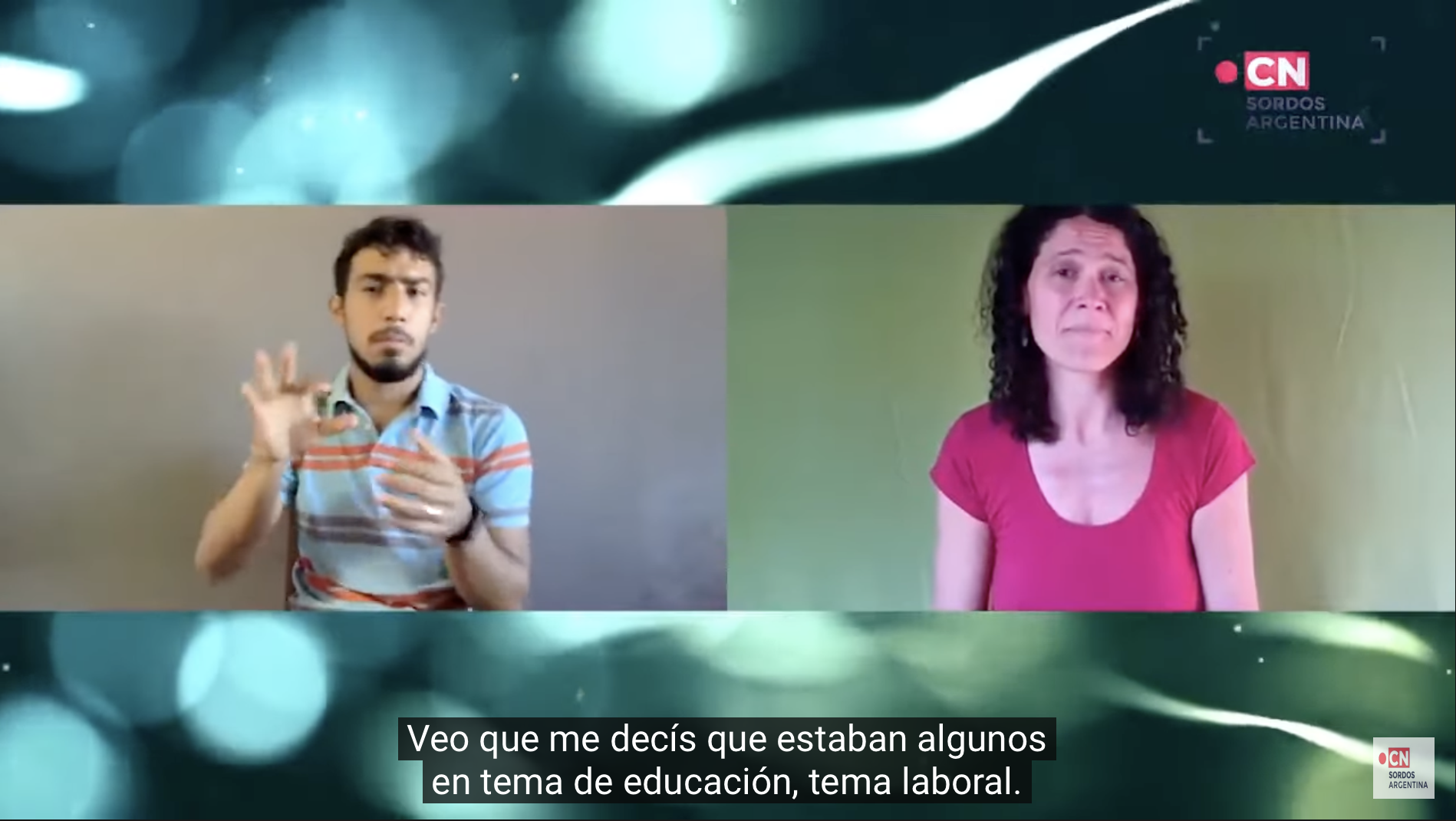}
         \caption{"Ley federal LSA"}
         \label{subfig:ley_federal_lsa}
     \end{subfigure}
     \hfill
     \begin{subfigure}[b]{0.47\textwidth}
         \centering
         \includegraphics[width=\textwidth]{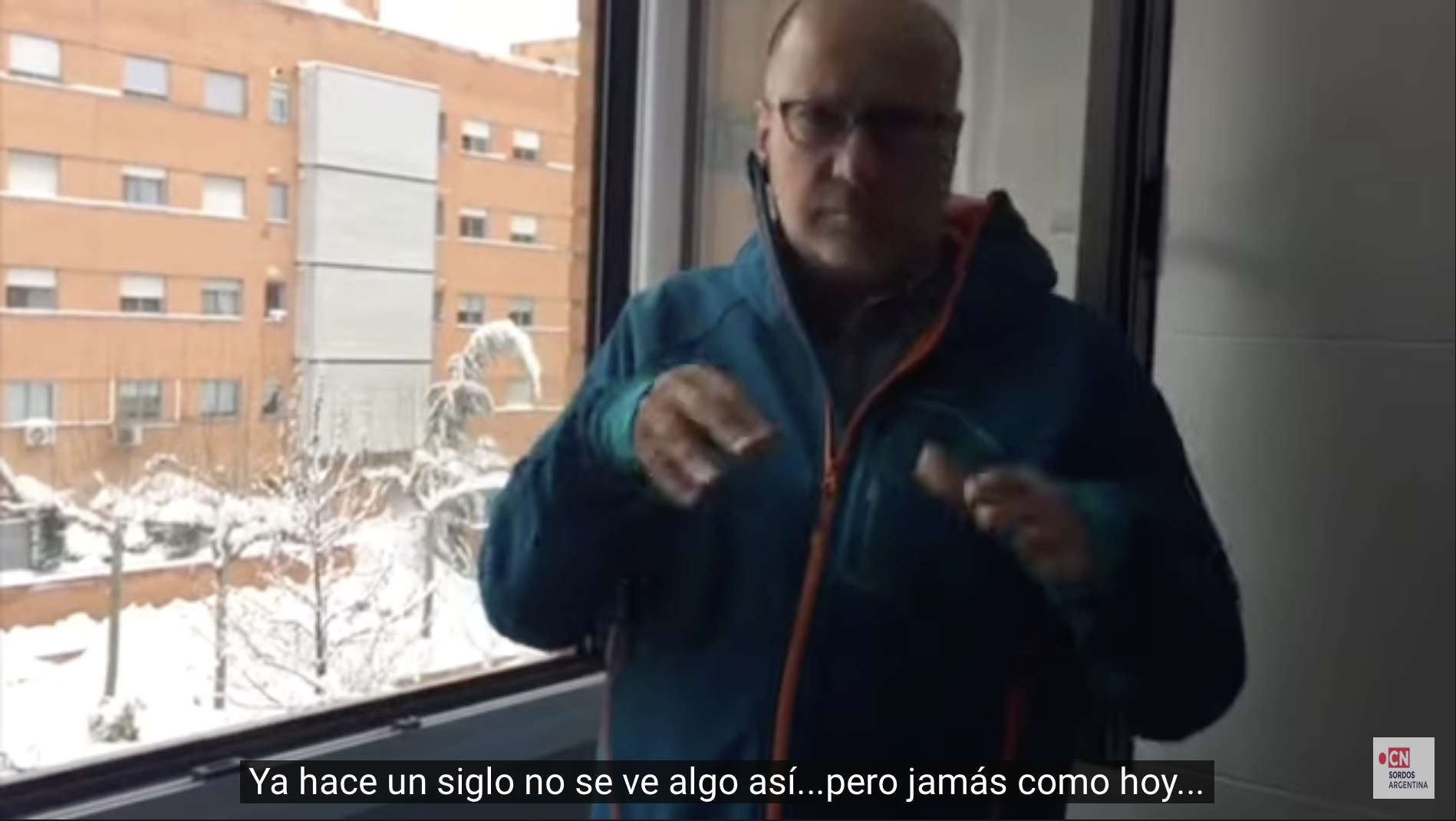}
         \caption{"Último momento"}
         \label{subfig:ultimo_momento}
     \end{subfigure}
     \hfill
     \begin{subfigure}[b]{0.47\textwidth}
         \centering
         \includegraphics[width=\textwidth]{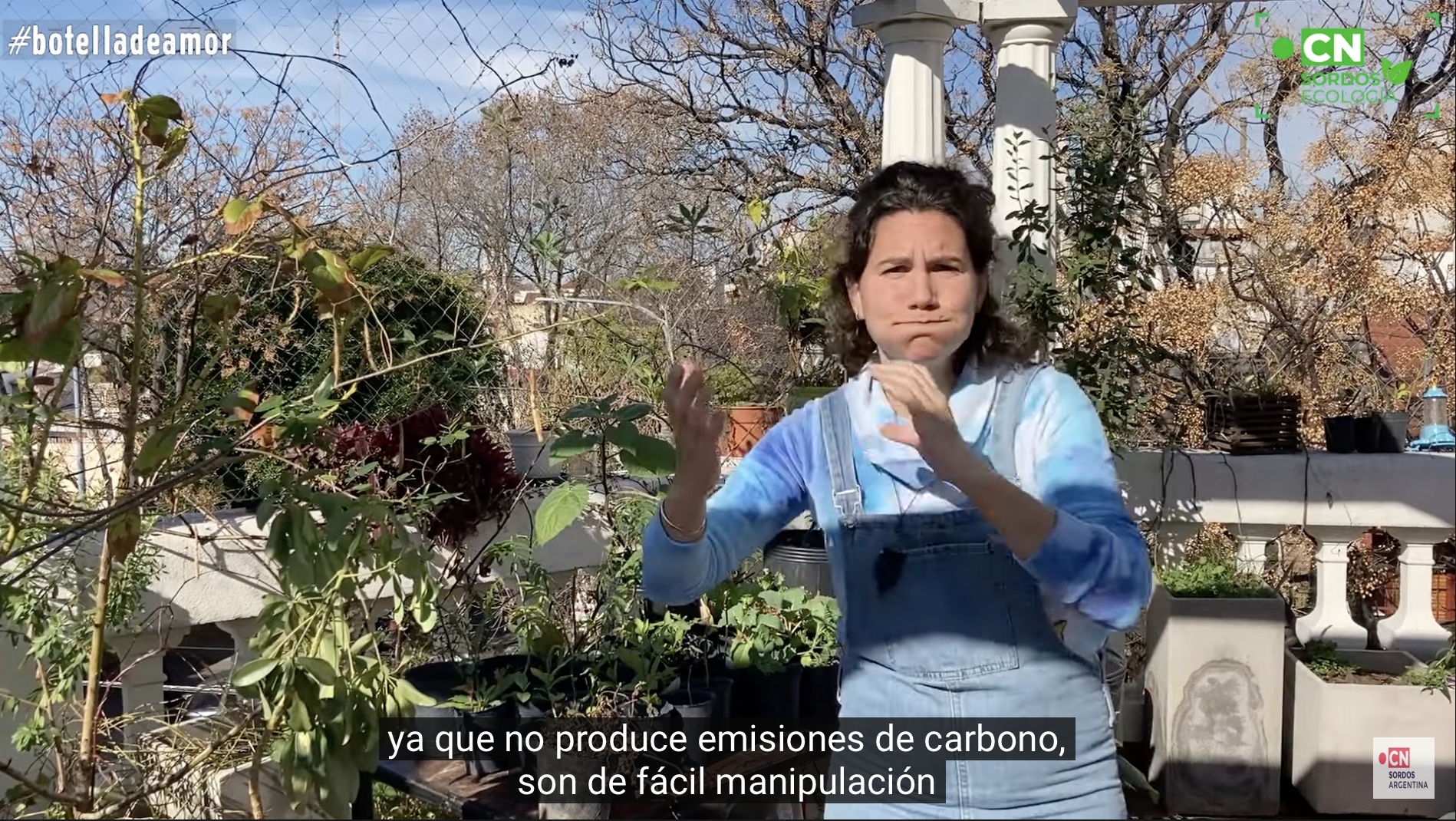}
         \caption{"CN Sordos ecología"}
         \label{subfig:cn_sordos_ecología}
     \end{subfigure}
    \caption{Screenshots of videos from the different channel playlists.}
    \label{fig:channel_screenshots}
\end{figure}

Because of the aforementioned, a dataset that allows training models for video translation from LSA to text (and also generating LSA videos from text, altought not the focus of this work) is thought to be especially useful in order to improve the communication and inclusion of deaf people.

\subsection{Contributions}

The main contribution of this work is the presentation and analysis of \dbname: the first continuous LSA dataset. The dataset contains:

\begin{itemize}
    
    \item More than 20 hours of video extracted from CN Sordos \cite{url:cnsordos}, a YouTube channel that presents the latest news in LSA.
    
    \item The corresponding Spanish translation for the sign sentences in each video.
    
    \item The keypoints for each signer, computed using AlphaPose \cite{fang2017rmpe}.
    
    \item An estimation of the active signer in the videos where multiple people appear.

\end{itemize}

Figure \ref{fig:channel_screenshots} shows images of some of its videos. We also present a visualization tool for \dbname that allows to explore the videos present in the dataset. Finally, a model for SLT trained on this dataset is presented as a baseline. The model uses the keypoints to infer the corresponding Spanish sentence. Details of the model and its resulting metrics are described in section \ref{sec:experiments}.

\section{Related work}
\label{sec:related}
Sign language datasets can be categorized into isolated or word-level, where each video matches a specific sign, and continuous or sentence-level, where videos contain many signs that correspond to an entire sentence. As mentioned before, while isolated datasets can be used for recognition they cannot be used for translation \cite{bragg2019sign}. Since the main goal of this work is to present a dataset for SLT, only CSL datasets will be considered.

One of the most relevant SLT datasets is RWTH-Phoenix-Weather 2014 T \cite{camgoz2018neural}. It contains videos of German Sign Language (GSL) extracted from German public TV weather forecasts. This dataset is used today as the main benchmark for SLT and, having a vocabulary of over 1000 signs, is considered the only resource for large-scale continuous sign language worldwide \cite{koller2020quantitative}. As the present work, it has the peculiarity of having been recorded in real life conditions, which may result in a more challenging dataset than if it was laboratory-made. Most of the other available SLT datasets have been recorded in laboratory conditions and contain a set of frequent or relevant sentences for its respective language. This is the case of the SIGNUM dataset \cite{von2008significance} of GSL, the Chinese Sign Language dataset (CSL) \cite{huang2018video}, the Greek Sign Language (GSL) dataset \cite{adaloglou2020comprehensive} and the KETI dataset \cite{ko2019neural} of Korean Sign Language.

Table \ref{tab:bd_comparison}  shows the main features of the mentioned datasets alongside \dbname. Many datasets also provide information about glosses translation (an intermediate translation between sign language and spoken language where each gloss corresponds to one sign) so we differentiated between gloss (gl) and word (w) level. There is no information about glosses in \dbname because it only contains sentence-level Spanish translation. There is no gloss information about CLS dataset either, as there is no gloss notation in CSL, signs map directly to words.

\begin{table}
\caption{Main features of the studied datasets.}
\label{tab:bd_comparison}
\begin{center}
    \begin{tabular}{ |l|c|c|c|c|c|c| }
        \hline
        & PHOENIX* & SIGNUM & CSL & GSL & KETI & \dbname\\ 
        \hline
        language & German & German & Chinese & Greek & Korean & Spanish\\
        sign language & GSL & GSL & CSL & GSL & KLS & LSA\\
        real life & \textbf{Yes} & No & No & No & No & \textbf{Yes}\\
        signers & 9 & 25 & 50 & 7 & 14 & \textbf{103}\\
        duration [h] & 10.71 & 55.3 & \textbf{100+} & 9.51 & 28 & 21.78\\ 
        \# samples & 7096 & \textbf{33,210} & 25,000 & 10,295 & 14,672 & 14,880\\ 
        \hline
        \# unique sentences & 5672 & 780 & 100 & 331 & 105 & \textbf{14,254}\\
        \% unique sentences & 79.93\% & 2.35\% & 0.4\% & 3.21\% & 0.71\% & \textbf{95.79\%}\\
        vocab. size (w) & 2887 & N/A & 178 & N/A & 419 & \textbf{14,239}\\ 
        \# singletons (w) & 1077 & 0 & 0 & 0 & 0 & \textbf{7150}\\
        \% singletons (w) & 37.3\% & 0\% & 0\% & 0\% & 0\% & \textbf{50.21\%}\\
        vocab. size (gl) & \textbf{1066} & 450 & - & 310 & 524 & -\\
        \# singletons (gl) & \textbf{337} & 0 & - & 0 & 0 & -\\
        \# singletons (gl) & \textbf{31.61\%} & 0\% & - & 0\% & 0\% & -\\
        \hline
        resolution & 210x260 & 776x578 & \textbf{1920x1080} & 848x480 & \textbf{1920x1080} & \textbf{1920x1080}\\
        fps & 25 & \textbf{30} & \textbf{30} & \textbf{30} & \textbf{30} & \textbf{30}\\
        \hline
    \end{tabular}
*Data was not available for the whole PHOENIX dataset, so the table show its train set statistics.
\end{center}
\end{table}

It is interesting to notice that there is a significant difference between datasets generated in real life conditions and those generated in laboratory conditions regarding the amount of unique sentences, the vocabulary size and the amount of singletons (tokens that appear only once in the training set). As laboratory-made dataset first choose a number of sentences and then record them many times, the set of unique sentences is a small fraction of the total amount of samples (between 0 and 5 percent). On the other hand, as it is not as common to find an exact sentence repeated many times in realistic environments, datasets generated in real life conditions present a much higher rate of unique sentences: 79.93\% for the PHOENIX dataset and 95.79\% for LSA-T. This rate was expected to be lower in PHOENIX than in LSA-T as the videos in PHOENIX were taken only from the weather forecast, while videos in LSA-T feature a wide range of topics. This directly impacts the size of the vocabulary and the amount of unique words in each dataset. While laboratory-made datasets have a small vocabulary with no singletons (all sentences appear multiple times), datasets collected in real-life environments have large vocabulary sizes with a significant amount of unique words (37.3\% in PHOENIX and 50.21\% in LSA-T). As a consequence, datasets recorded in real-life conditions, and particularly the one presented in this work, are highly challenging for translation models.

\subsection{LSA}

LSA is the sign language used by the deaf community in Argentina. In 2018, 0.02\% of the Argentinian population older than 6 years old presented a hearing difficulty, half of them being deaf \cite{instituto2018estudio}. Counting teachers, relatives and friends of deaf people, there are 2 million LSA users \cite{url:lsa}.

Regarding LSA datasets, LSA64 \cite{ronchetti2016sign}\cite{ronchetti2016lsa64} is, to the best of our knowledge, the only other dataset available. It contains 3200 videos of 64 isolated signs recorded by a team of researchers from the Universidad Nacional de La Plata. As it has been generated under laboratory conditions, the background scenarios are similar. Although it has been the only source of data for training LSA sign recognition models, it cannot be used for translation.

\subsection{SLT}

Regarding SLT, works that use keypoints data as part of the input for a translation model can be found in \cite{borg2020phonologically} and  \cite{zhou2020spatial}. In those cases keypoints data were given to the model as extra information together with the corresponding video. Furthermore, the models presented in \cite{ko2019neural} and \cite{kim2022keypoint} have a similar approach to the one presented in this work. Also, they only use keypoints information as input for a neural network model. In both cases the authors train a recurrent network on the KETI dataset. Particularly, in \cite{kim2022keypoint} no gloss information is employed, neither as input or output.

\section{\dbname dataset}
\label{sec:dataset}
\dbname was built from videos from the YouTube channel CN Sordos, a news channel created in 2020 by deaf people and deaf people's relatives. The hosts use LSA to communicate the news. Subtitles in Spanish are provided by the authors. There are two main hosts in most of the videos. There are other regular hosts in charge of special sections like ecology or movies, but these appear much less often. Therefore, the great majority of the 103 signers present on the dataset are guests who appear once or twice for an interview or a specific piece of news. There is gender parity among the signers and videos contain different locations, backgrounds and lighting conditions.

\dbname was generated in two stages. In the first stage, we collected the videos and cropped them according to its subtitles. For each valid sentence in the subtitles of a video (annotations like "[background music]" were discarded), we cropped the corresponding video clip. This process resulted in 14,880 video segments. In the second stage, in those segments where there is more than one person, we inferred who is the signer with a certain score of confidence. Section \ref{subsec:signer_inference} describes how this inference was done. 

Finally, we also made available a visualization tool so that researchers can easily explore the dataset. The tool was developed using Fifty One \cite{moore2020fiftyone}, which provides useful features such as allowing to filter samples by label, video, playlist, or by the confidence score of the signer inference.

Full code for the development of this dataset, the analysis described below, and the start-up of the visualization tool can be found in the project's repository\footnote{\dbrepourl}.

\subsection{Statistics and analysis}\label{subsec:stats}

\begin{figure}
    \centering
    \begin{subfigure}[b]{0.38\textwidth}
        \centering
        \includegraphics[width=\textwidth]{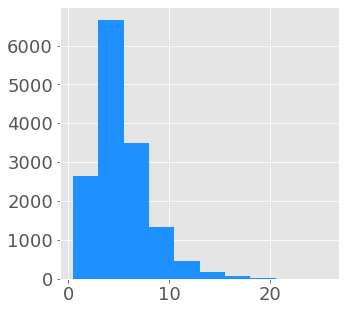}
        \caption{}
        \label{subfig:hist_times_per_vid}
    \end{subfigure}
    \hfill
    \begin{subfigure}[b]{0.38\textwidth}
        \centering
        \includegraphics[width=\textwidth]{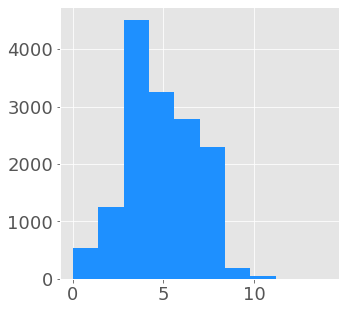}
        \caption{}
        \label{subfig:hist_words_per_vid}
    \end{subfigure}
    \hfill
    \caption{Histograms of segments duration (a) and amount of words per sentence (b).}
    \label{fig:clips_statistics}
\end{figure}

\dbname was built from 63 different videos of the CN Sordos Channel. We only used videos that had been annotated with subtitles by the authors (LSA experts). The videos resulted in 14,880 segments that span 21.78 hours of video. Each sample of the database contains a sentence taken from the subtitles provided and its respective video segment. The generated segments have a mean duration of 5.27 seconds and their labels a mean of 4.78 words per sample (Figure \ref{fig:clips_statistics}).

\begin{figure}
    \centering
    \begin{subfigure}[b]{0.32\textwidth}
        \centering
        \includegraphics[width=\textwidth]{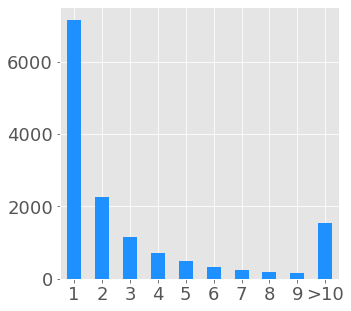}
        \caption{}
        \label{subfig:hist_word_freq}
    \end{subfigure}
    \hfill
    \begin{subfigure}[b]{0.32\textwidth}
        \centering
        \includegraphics[width=\textwidth]{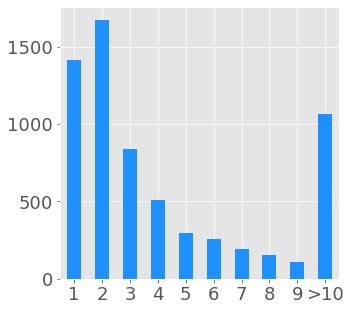}
        \caption{}
        \label{subfig:hist_words_freq_no_sing}
    \end{subfigure}
    \hfill
    \begin{subfigure}[b]{0.32\textwidth}
        \centering
        \includegraphics[width=\textwidth]{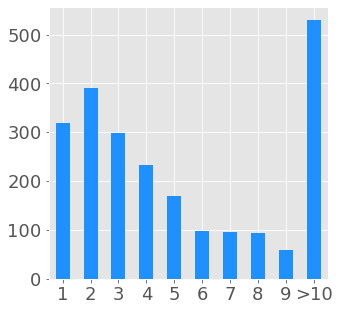}
        \caption{}
        \label{subfig:hist_words_freq_no_lt5}
    \end{subfigure}
    \caption{Histograms of word frequency over the whole dataset (a), after removing samples with singletons (b) and and after removing samples with words with frequency lower than 5 (c).}
    \label{fig:words_statistics}
\end{figure}

As shown in Table \ref{tab:bd_comparison}, a peculiarity of the dataset is the significant proportion of unique sentences (\(\sim95\%\)) and the proportion of unique words with respect to the vocabulary size (\(\sim50\%\)) (\ref{subfig:hist_word_freq}). Therefore, we consider the dataset to be highly challenging for a SLT model. 

A common preprocessing step in the natural language processing field consist on removing sentences with words that appear very few times in the dataset so they do not add noise to the training process. Histograms of word frequencies after removing samples with singletons and words with frequencies lower than 5 are shown in figures \ref{subfig:hist_words_freq_no_sing} and  \ref{subfig:hist_words_freq_no_lt5}, respectively. After removing the corresponding subset of labels, the histograms do not show such a heavy left tail, despite the fact that other words now appear with a lower frequency than before.

\begin{figure}
    \centering
    \begin{subfigure}[b]{0.4025\textwidth}
        \centering
        \includegraphics[width=\textwidth]{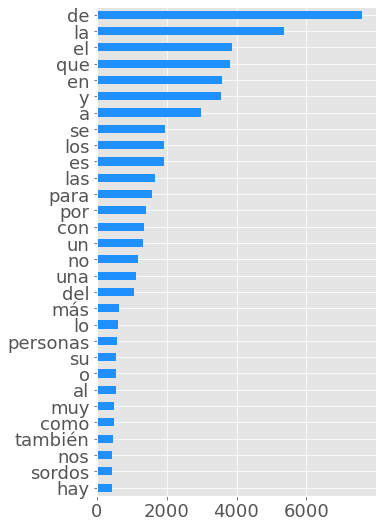}
        \caption{}
        \label{subfig:bar_30_most_common_words_freq}
    \end{subfigure}
    \hfill
    \begin{subfigure}[b]{0.57\textwidth}
        \centering
        \includegraphics[width=\textwidth]{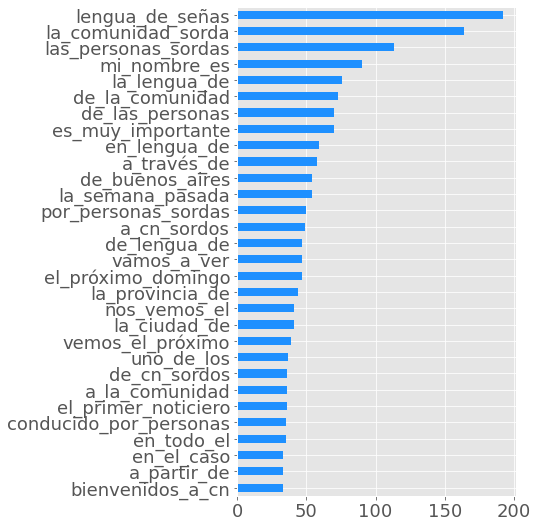}
        \caption{}
        \label{subfig:bar_30_most_common_trigrams_freq}
    \end{subfigure}
    \hfill
    \caption{Most common words (a) and trigrams (b) in the dataset}
    \label{fig:most_common_words_trigrams}
\end{figure}

Finally, figures \ref{subfig:bar_30_most_common_words_freq} and \ref{subfig:bar_30_most_common_trigrams_freq} show the 30 most commons words and trigrams respectively. As many of the most common words correspond to articles or connectors we believe the trigrams can provide a more representative visualization of the dataset content. Additionally, we expect traditional models without special training schemes to be able to learn these common trigrams.

\subsection{Default Train-Test sets}\label{subsec:train_test_split}

Alongside the dataset, we provide two versions of the train and test sets: a full version and a reduced version. We made the reduced version by excluding labels that contain words with frequency lower than 5. This version is expected to be easier for a SLT model to learn. In both versions we used 80\% of the samples for training and 20\% for testing. Samples with a signer-confidence score lower than 0.5 were discarded (section \ref{subsec:signer_inference}).

Also, as videos labels are subtitles instead of classes, traditional stratified Train-Test splits cannot be performed. Since samples from the same video are expected to share topics and therefore vocabulary, both train and test sets were generated so that segments from the same video are included in both.

\begin{table}
\caption{Comparison between the two versions of the train-test sets.}
\label{tab:train_test_comparison}
\begin{center}
    \begin{tabular}{ |p{4cm}|c|c c|c c| }
        \hline
        & \dbname & \multicolumn{2}{|c|}{Full version} & \multicolumn{2}{|c|}{Reduced version}\\ 
        & & Train & Test & Train & Test \\
        \hline
        \# sentences & 14,880 & 11,065 & 2735 & 3767 & 910\\
        \% unique sentences & 95.79\% & 96.64\% & 92.78\% & 96.88\% & 98.35\%\\
        \hline
        vocab. size & 14,239 & 12,385 & 5546 & 2694 & 1579\\
        \% singletons & 50.21\% & 52.01\% & 61.9\% & 23.2\% & 48.83\%\\
        \% sentences with singletons & 34.97\% & 40.98\% & 67.97\% & 14.36\% & 54.29\%\\
        \% sentences with words not in train vocabulary & - & - & 59.2\% & - & 84.5\%\\
        \hline
    \end{tabular}
\end{center}
\end{table}

Table \ref{tab:train_test_comparison} shows a comparison between the different versions. As it can be seen, there is no significant difference between them in terms of percentage of unique sentences, but there is in terms of percentage of singletons. While in the full version around 50\% of the training vocabulary are singletons, only 23\% of the train vocabulary of the reduced version are. Furthermore, while 40\% of the labels on the full version train train contain singletons, only 14\% of the labels of the reduced version train do. Apart from this, it is also worth noticing that, in both versions, there is an important amount of sentences (\(\sim60\%\) and \(\sim85\%\) respectively) of the test set that contains tokens that were not present in the train set. This must be considered when evaluating a SLT model over the dataset as it wont be able to learn about those tokens and its usage from the train set.

\subsection{Signer inference}
\label{subsec:signer_inference}

\begin{figure}
     \centering
     \begin{subfigure}[b]{0.4\textwidth}
         \centering
         \includegraphics[width=\textwidth]{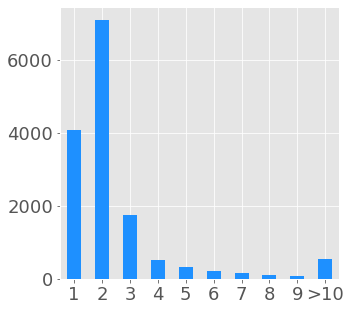}
         \caption{Histogram of amount of people detected by AlphaPose.}
         \label{subfig:amount_of_people_hist}
     \end{subfigure}
     \hfill
     \begin{subfigure}[b]{0.4\textwidth}
         \centering
         \includegraphics[width=\textwidth]{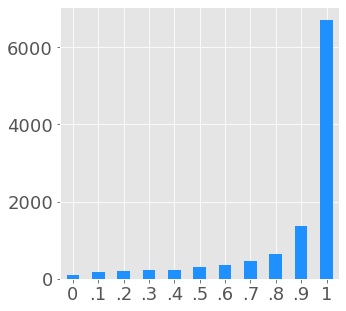}
         \caption{Histogram of signer-confidence scores (rounded to one decimal).}
         \label{subfig:signers_scores_hist}
     \end{subfigure}
    \caption{Amount of signers and signer scores histograms.}
    \label{fig:signers_statistics}
\end{figure}

Many videos in \dbname feature more than one person. In most cases, they appear at the same distance from the camera and in similar positions (Figure \ref{fig:channel_screenshots}), so identifying who is the active signer is not trivial. Figure \ref{subfig:amount_of_people_hist} presents an histogram of samples by number of people detected. Most of the videos have two people in them and only a third of the samples feature one person. Identifying the signer is particularly useful as it allows to define a region of interest (ROI) that models could focus on while performing SLT.

To infer the signer, we first obtained the keypoints and bounding boxes of the people in each video. This was done using AlphaPose with the Halpe full-body keypoints format \cite{fang2017rmpe}. This format has 136 keypoints of which 42 are hand keypoints, which makes them suitable for hand gesture recognition. Then, we transformed each keypoints' coordinates to a relative position with respect to the person's center. For each person in the video, we computed the total amount of movement of their hands. The person chosen as the signer is the one who has the largest amount of hand movement, with a high confidence (Figure \ref{subfig:signers_scores_hist}). 

The resulting keypoints computed by AlphaPose, the result of the signer inference and the amount of movement of each person used to compute the confidence score of the inference are provided for each sample of the dataset.

\section{Baseline experiments} 
\label{sec:experiments}
\subsection{Proposed approach}

A SLT model trained on \dbname is proposed to serve as baseline for future comparisons. The model used is similar to the one presented in \cite{kim2022keypoint}, using only the keypoint data extracted from the videos as input. However, instead of using recurrent layers we use a transformers as the main building blocks of the network.


The proposed model then takes as input a list of \(F\) frames and predicts the words of its corresponding sentence one at a time. Each frame is a matrix of size \(2K\) that has the \(x\) and \(y\) position for the \(K\) keypoints and each sentence is encoded as a list of the embeddings of its words.

\subsection{Metrics}

Given the difficulty of \dbname dataset caused by having many words that appear few times, we proposed a new metric called WER-N, based on the traditional Word Error Rate (WER). WER-N mimics the top-N accuracy for classification models by not penalizing the model when it misses words from the target that were seen less than \(N\) times in the training set. The metric can be used in any problem that present tokens with low frequency.

WER-N can be computed using a variant of the Edit Distance algorithm. For two strings \(s\) and \(t\), the distance \(d\) between the sub-strings \(s[0:i]\) and \(t[0:j]\) can be computed recursively using the formula:

\begin{equation}
\begin{split}
    d[i,j] = min( &
       d[i-1,j] + C(I)\\
       & d[i,j-1] + C(D^*)\\
       & d[i-1,j-1] + C(S^*))
\end{split}
\end{equation}

Where \(C(I)\), \(C(D^*)\) and \(C(S^*)\) represent the cost of an insertion, deletion and substitution respectively. The WER-N metric modifies the traditional definitions of C(D) and C(S) by not penalizing some errors as follows.:

\begin{equation}
\begin{split}
    C(D^*) & =
        \begin{cases}
            0 & \text{if frequency of token } t[j] \text{ is lower than } N \\
            1 & \text{otherwise}
        \end{cases}\\
    C(S^*) & =
        \begin{cases}
            0 & \text{if } s[i]=t[j]\\
            0 & \text{if frequency of token } t[j] \text{ is lower than } N \\
            1 & \text{otherwise}
        \end{cases}
\end{split}
\end{equation}

\subsection{Experimental results}

We performed two sets of experiments, using the two versions of the train and test sets presented in section \ref{subsec:stats}. We implemented the proposed model using PyTorch 
and trained the model for 30 epochs in each experiment. We used the Adam optimizer
and a learning rate of 0.0001. We used 75 frames per sample, obtaining an average of \(\sim15\) FPS over all samples in the dataset.

The size of both the keypoint embeddings and the word embeddings was set to 64. The parameters for defining the transformer network were the same as the proposed in the original paper \cite{vaswani2017attention}. We used the Spacy Spanish \cite{spacy2} tokenizer for the sample labels.

Only the 42 hands keypoints were used as the model input as those were the ones considered relevant for SLT. Finally, we used a threshold of \(T=0.3\) for the keypoint confidence. Full code for the model development, training and experimentation can be found in its GitHub repository\footnote{\modelrepourl}.

\begin{table}
\caption{Metrics for the proposed model over the two train-test sets}
\label{tab:model_metrics}
\begin{center}
    \begin{tabular}{ |l|c c|c c| }
        \hline
        & \multicolumn{2}{|c|}{Full version} & \multicolumn{2}{|c|}{Reduced version}\\ 
        & Train & Test & Train & Test\\
        \hline
        WER & 0.9387 & 0.9392 & 0.9207 & 0.957\\
        WER-5 & 0.8116 & 0.7892 & 0.7982 & 0.68\\
        WER-10 & 0.7547 & 0.7154 & 0.7193 & 0.5904\\
        \hline
    \end{tabular}
\end{center}
\end{table}

Table \ref{tab:model_metrics} shows the result of evaluating the model over the two versions of the train-test sets. We used WER and the proposed WER-N with \(N=5\) and \(N=10\) as evaluation metrics. As it can be seen, the model does not perform well, although similar approaches had obtained decent results in other datasets like \cite{ko2019neural} or \cite{kim2022keypoint} on KETI dataset. It is interesting to notice that \cite{kim2022keypoint} also tried the same model over RWTH Phoenix dataset, that as table \ref{tab:bd_comparison} showed, has a bigger similarity to ours than the laboratory made datasets as KETI. Their model achieved a BLEU score of \(\sim85\) over KETI, but when evaluated over Phoenix it obtained a BLEU score of \(\sim13\) across all their experiments. Then it is not surprising that given the complexity of the presented dataset the models did not obtained good results.

\section{Conclusions and Future work}
\label{sec:conclusions}
We have presented a new dataset from the Argentinian Sign Language. It is, to the best of our knowlegde, the first continuous dataset that will make it possible to train SLT models for LSA.

The dataset was built using videos from the channel CN Sordos and shows a wide range of topics, signers, backgrounds and lightning conditions. It contains clips extracted from the videos, their corresponding label, the keypoints for each person in the video and, in case there is more than one person, an estimation of who is the one signing. Also a visualization tool was made available.

We have also proposed two possible versions of train and test sets and trained on them a translation model that uses keypoint information to serve as baseline.

As future works, we intend to train different state-of-the-art models over the dataset and compare their performance with the one achieved in other datasets such as Phoenix-RWTH. This models could take both video and keypoints information. Also, we plan on using data augmentation and one-shot learning techniques that will allow us to extract more information from the tokens or sentences that seldom appear on the dataset. Finally, we are working on a web tool that will allow people to use the translation model over a video recorded by themselves using a web cam.

%
%
%
\bibliographystyle{splncs04}
\bibliography{bibliography}

\end{document}